\documentclass[UTF]{sigkddExp}

\begin{document}
%

\title{Predictions of short-term driving intention using recurrent neural network on sequential data}
%

\numberofauthors{1}

%


\author{
%
\alignauthor Zhou Xing, Fei Xiao\\
       \affaddr{Athena Robotics, INC.}\\
       \affaddr{1999 S Bascom Ave}\\
       \affaddr{San Jose, CA 95008}\\
       {joe.xing.ai, feixiao}@gmail.com \\
}
\date{11 January 2017}

\maketitle

\begin{abstract}

Predictions of driver's intentions and their behaviors using the road is of great importance for planning and decision making processes of autonomous driving vehicles. In particular, relatively short-term driving intentions are the fundamental units that constitute more sophisticated driving goals, behaviors, such as overtaking the slow vehicle in front, exit or merge onto a high way, etc. While it is not uncommon that most of the time human driver can rationalize, in advance, various on-road behaviors, intentions, as well as the associated risks, aggressiveness, reciprocity characteristics, etc., such reasoning skills can be challenging and difficult for an autonomous driving system to learn. In this article, we demonstrate a disciplined methodology that can be used to build and train a predictive drive system, therefore to learn the on-road characteristics aforementioned.

\end{abstract}

\section{Introduction}

Predictions of driver intentions and their on-road behaviors is of great importance for a safe and defensive path planning and decision making system of autonomous driving vehicles. In particular, short-term driving intentions are the fundamental building blocks consisting of relatively long-term and more sophisticated goals, such as overtaking the slow vehicle in front, taking an exit or merge onto a high way with condensed traffic, etc. While it is not uncommon that most of the time human driver can predict, few seconds in advance, simple intentions of other drivers,  various on-road behaviors, thus rationalize the associated risks, such reasoning capabilities can be challenging and difficult for an autonomous driving system to be equipped with.

In this article, we present a disciplined methodology to build and train a predictive drive system, which includes various components such as traffic data, traffic scene generator, simulation and experimentation platform, supervised learning framework for sequential data using recurrent neural network (RNN) approach, validation of the modeling using both quantitative and qualitative methods, etc. In particular, the simulation environment, in which we can parameterize and configure relatively challenging traffic scenes, customize different vehicle physics and controls for various types of vehicles such as cars, SUV, trucks, test and utilize high definition map of the road model in algorithms, generate sensor data out of light detection and ranging (LIDAR), optical wavelength cameras for training deep neural networks, is crucial for driving intention, behavior, collision risk modeling since collecting statistically significant amount of such data as well as experimentation processes in the real world can be extremely time and resource consuming. It is our intention that standardizing such a testing, scoring system can be very useful to validate and experiment various planning and prediction algorithms of autonomous driving application.

Recurrent neural networks can be used to learn features and long term dependencies from sequential time-series data. The general RNN architecture have a stack of non-linear activation components populated in the latent space from low to high dimensionality, where some of the components, denoted as ``cell", form a recurrence structure along the time dimension which allows temporal correlations being conveyed throughout the sequence. We experiment with the different recurrent structures, such as Long short-term memory (LSTM \cite{DBLP:journals/corr/Graves13, DBLP:journals/corr/SutskeverVL14}, gated recurrent unit (GRU) \cite{DBLP:journals/corr/ChungGCB14, DBLP:journals/corr/ChoMBB14}, to predict the sequence intentions, where, instead of projecting the cell hidden states to future for sequence generation, we reformalize the use as a classification problem similar to other classification tasks such as image, music classfications \cite{NIPS2012_4824, DBLP:journals/corr/SzegedyLJSRAEVR14, DBLP:journals/corr/XingBJKLMPRSXP17}. Subtleties of this classification problem reside in taking into account the temporal correlations within the sequence, i.e., classifying the intention labels at each time step of a sequence is necessary for a better capture of the intention transitions. We implement time dependent classifications by aggregating and comparing the output state vector of the recurrent cell in the RNN with the training labels at each time step of a sequence as shown in Fig.~\ref{fig:rnn}. The results of our approach are reported in Section \ref{sec:quantitative_measurement} and \ref{sec:qualitative_measurement}.

\section{Description of the simulation environment and data format} 

\begin{figure}[!ht]
  \centering
    \includegraphics[width=0.5\textwidth]{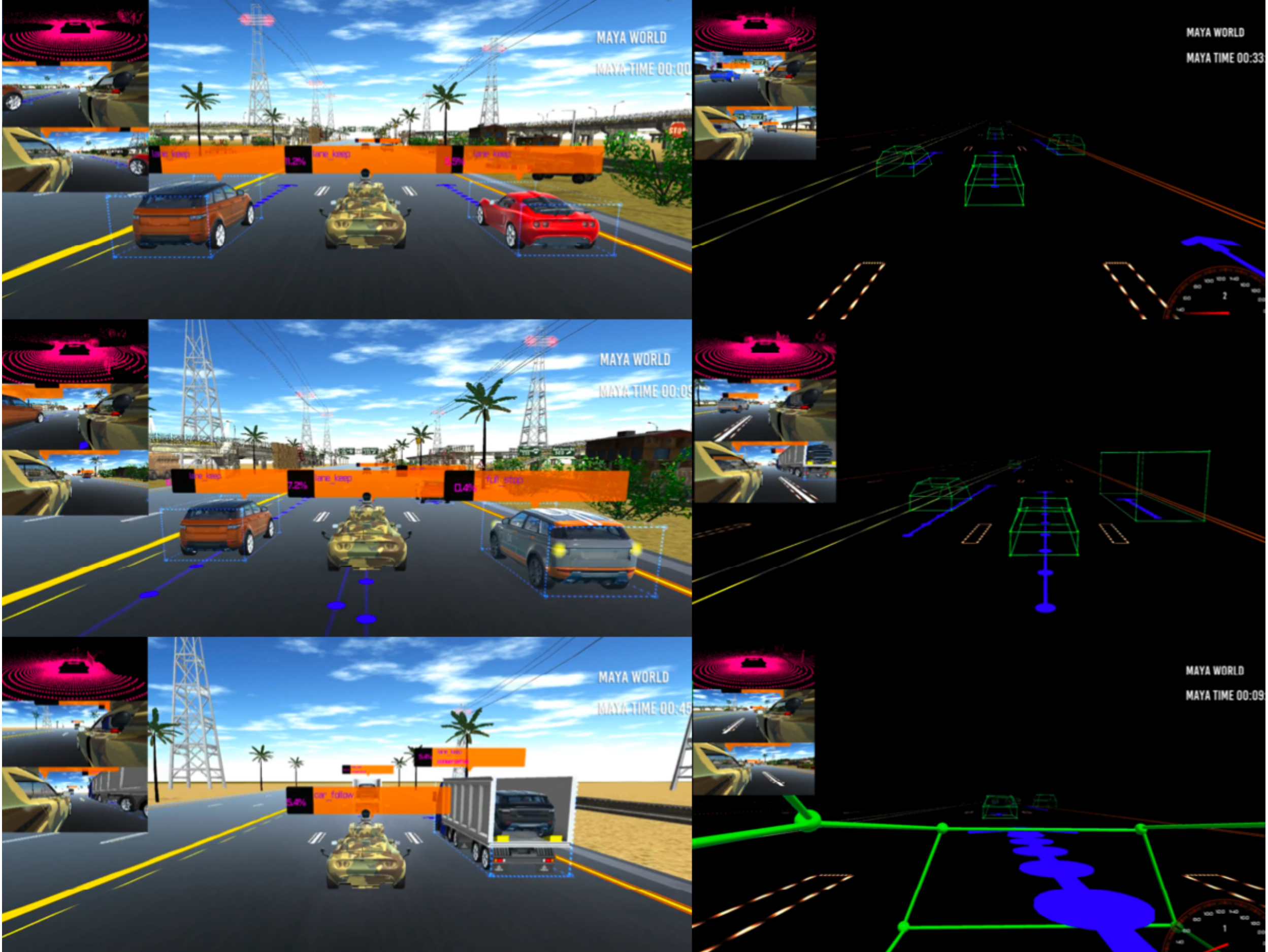}
\caption{Basic maneuvers such as lane keep, car following, stopping, etc., are illustrated in the high way scene. The left side shows the rendering of the simulated scene while the right hand side shows the ``skeleton" version of the scene. }
\label{fig:lane_keep}
\end{figure}

\begin{figure}[!ht]
  \centering
    \includegraphics[width=0.5\textwidth]{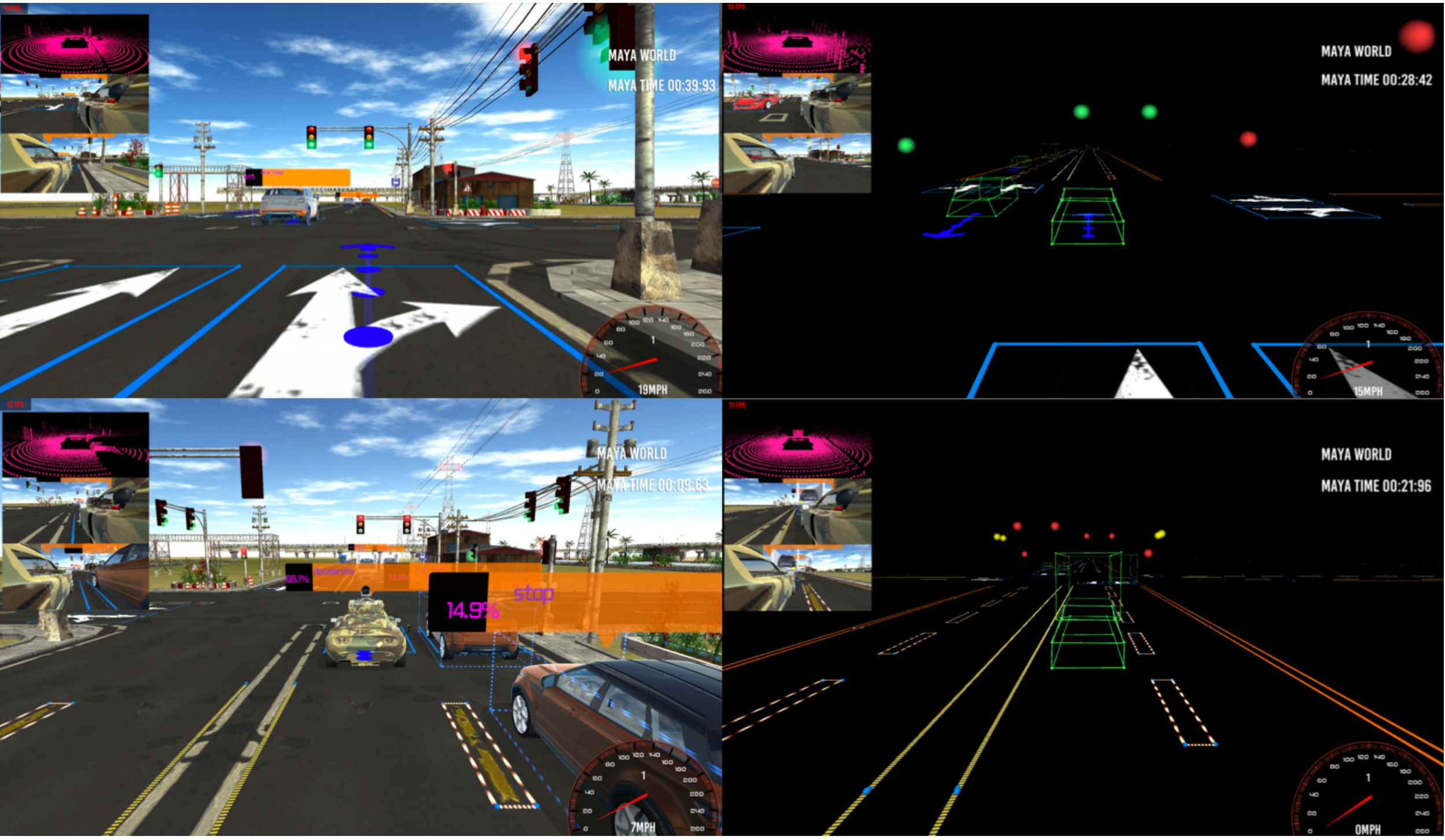}
\caption{Scene of urban driving is shown both in regular rendering and ``skeleton" mode with traffic lights, HD map information provided.}
\label{fig:urban}
\end{figure}

In order to experiment, validate our planning and prediction algorithms, we build a 3D simulator, denoted as ``maya", using Unity3D development suite \cite{Unity3D}. Various traffic scenes such as high way exit or entrance, making a left turn at intersections on urban streets when there is no explicit traffic light guidance (no left-turn green arrow), U-turns at intersection, are created for experimentations. We have opened some of these scenes to public access \cite{maya_docker} for research purposes as shown in Fig.~\ref{fig:lane_keep} and Fig.~\ref{fig:urban}.

The other main purpose of maya is to generate large volume of training data for our deep learning, reinforcement learning algorithms. Maya can generate basic format of autonomous driving vehicle data including LIDAR point cloud (top left sub-figure in Fig.~\ref{fig:lane_keep} and Fig.~\ref{fig:urban}), camera images (bottom two left sub-figure and the main figure in Fig.~\ref{fig:lane_keep} and Fig.~\ref{fig:urban}), as well as high definition (HD) map data of road, lane geometries, traffic light status, etc., and send them out either as offline training dataset or online processing or training for systems such as reinforcement learning.

The so called ``ego vehicle" is the car with camouflage color, which is equipped with a simulated LIDAR device \footnote{The black device with a base sitting on top of the car is the simulated LIDAR device, from which all the ray cast measurements are issued.}, as well as three cameras: two on the side, one on the dash board facing front. The LIDAR point cloud consist of a parametrizable number (default to $10, 000$ ) of measurements of 3D positions which are generated using the ``ray cast'' method in Unity3D, each point is a measurement of collision between a ray tracing from the center of the simulated LIDAR device to the surrounding objects such as road, trees, posts, opponent vehicles, so on the so forth. Image data is in format of RGB intensities with parametrizable number of pixels. HD map data includes features on the road surface such as road segment polygons, lane marker polygons, road boundaries, road divider positions, and the traffic light status. We also provide perceived information such as the 3D bounding box coordinates in both LIDAR point cloud and image, velocity, tire angle, control signals including throttle, brake, steering wheel, hand brake, left and right signal lights of each vehicle. All of these data are available through either inter-process communication (IPC) using shared memory mechanism, or TCP/IP socket communication. The simulation environment can be visualized in either normal graphics rendering mode or the so called ``skeleton" mode, which is dedicated for debugging the path planner, prediction algorithms. In skeleton mode, only the skeletons of the vehicles, HD map information, traffic lights are displayed in a light-weight manner as shown on the right-hand-side of Fig.~\ref{fig:lane_keep} and Fig.~\ref{fig:urban} where the blue dots and arrow stand for the trajectory generated by the path planner.

While the path planning routine that is used to generate the ``ground truth" training labels for intention predictions is described elsewhere in Sec.~\ref{sec:path_planning}, here we give a brief summary of the dataset that has been prepared for training the RNN as shown in Table.~\ref{tab:data_for_segmentation}. Four datasets are made available where dataset A and B are collected in the high way scene, dataset C and D are collected in the urban traffic scene. Two different sequence length of 12 frames and 6 frames are used with the same readout rate of 5 Hz (200 ms) for every frame. All the sequences are randomly collected with different seedings, some of the sequences contain a transition of intentions for example from lane keeping to lane-change, or vice versa, from lane keep to car follow, from car follow to decelerate, etc. The statistics of each sample are also listed in the table, where both training features and ground truth labels at each time step are saved together in Tensorflow TFRecords \cite{Tensorflow} using SequenceExample protocol buffer.

\begin{table}
\scalebox{0.7}{\parbox{.5\linewidth}{%
\begin{center}
    \begin{tabular}{ | c | l | l | l | l | l |}
    \hline
    Dataset & Description & Data readout  	       & 	     Number of                                   & \multicolumn{2}{|c|}{Number of } \\
    		 &                    & rate (Hz)                 &            frames per sequence                  & \multicolumn{2}{|c|}{sequences} \\
    \hline
    		\multicolumn{4}{|c|}{}								& train & test \\
    \hline
   A  &  Basic path 	           &   5   &   12   & $\sim$258k	&  $\sim$33k \\
        &  planning, high way  &       &           & 	&  \\
     \hline
    B  &  Basic path  		  &   5   &    6    &  $\sim$500k & $\sim$72k \\
        &  planning, high way  &      &       & 	 	       & \\
      \hline
    C  &   Basic path               & 5   & 12  & $\sim$206k & $\sim$83k \\
        &   planning, urban       &      &       & 	 	       & \\
    \hline
    D  &  Basic path               & 5   & 6    & $\sim$215k & $\sim$100k \\
        &   planning, urban      &      &       & 	 	       & \\
     \hline
     \end{tabular}
\end{center}
}}
\caption{The dataset that has been prepared for training and testing our intention prediction models.}
\label{tab:data_for_segmentation}
\end{table}

\section{Vehicle model and path planning}
\label{sec:path_planning}

We use Unity3D provided car physics, wheel physics, and a slip-based tire friction model to instantiate our vehicles, where characteristics such as engine torque, brake torque, wheel suspensions, slip, collisions, etc. are all taken into account and parametrizable to certain extent as shown in Table~\ref{tab:vehicle_physics}. Tire friction is considered for the wheel's forward (rolling) direction and sideways direction, respectively. In both directions it is first determined how much the tire is slipping (based on the speed difference between the tire's rubber and the road) and then this slip value is used to find out tire force exerted on the contact point.

\begin{table}
\scalebox{1}{\parbox{.5\linewidth}{%
\begin{center}
    \begin{tabular}{ | l | l |}
    \hline
    Parameter & Value \\
    \hline
    Engine torque & 590 lb-ft \\
    Brake torque & 1475 lb-ft $\times$ input ratio \\
    Hand krake Torque & 4425 lb-ft $\times$ input ratio \\
    Braking force & 15,000 N \\
    Number of gears & 6 \\
    Wheel / Tire model & slip based friction model \\
    Mass & 50 Kg\\
    Suspension distance & 0.3 m \\
    Suspension spring & 25,000 N\\
    Forward slip & stiffness or traction controlled \\
    Side way slip & stiffness or traction controlled \\
    Skid & slip limited\\
     \hline
     \end{tabular}
\end{center}
}}
\caption{The car and wheel physics that are configured as the baseline vehicle model in maya.}
\label{tab:vehicle_physics}
\end{table}

\begin{figure}[!ht]
  \centering
    \includegraphics[width=0.5\textwidth]{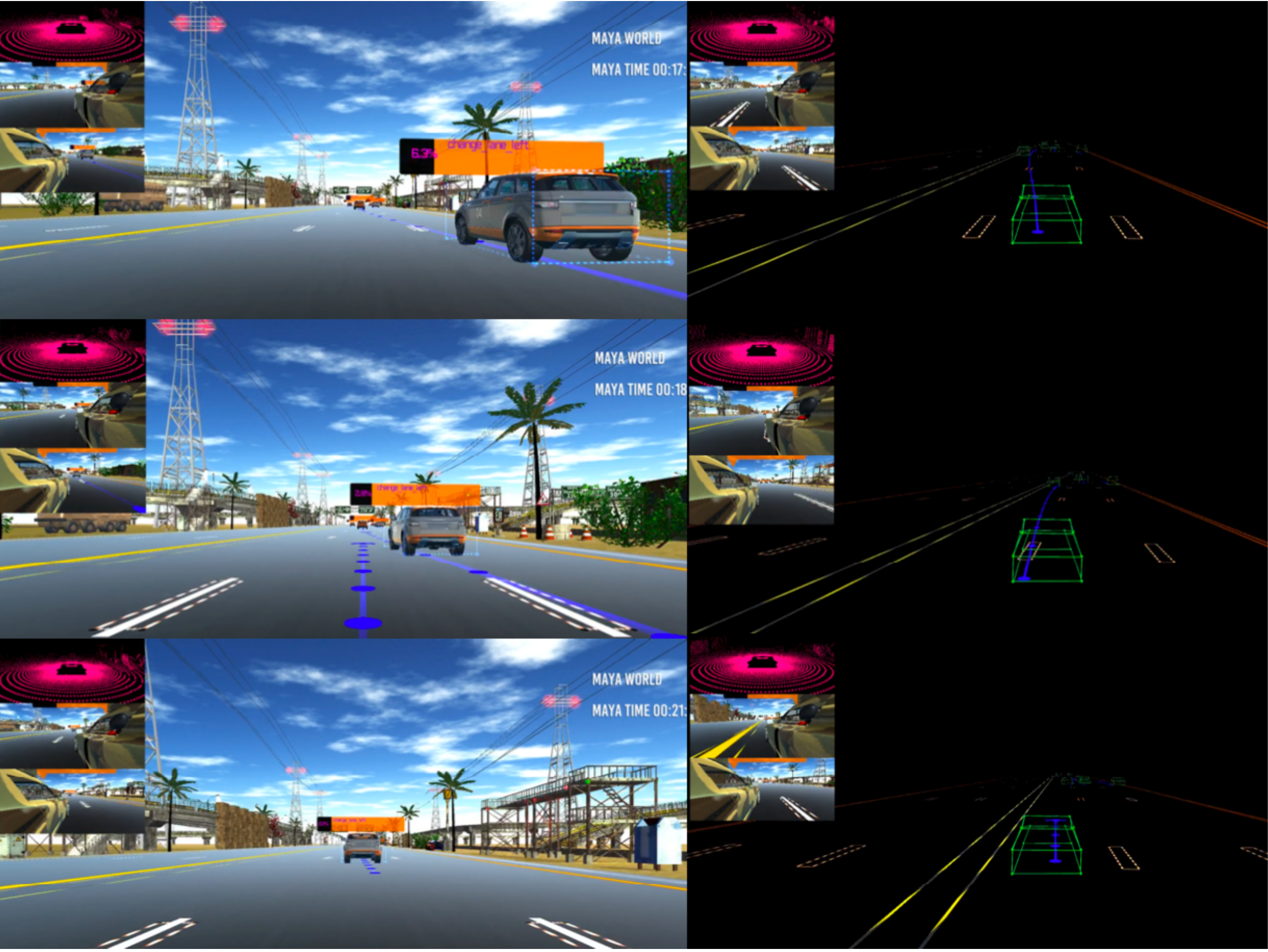}
\caption{Examples of intention of changing to the left lane as well as planned path is shown in the high way scene as the blue dots and arrow, where the left side shows the rendering of the simulated scene while the right hand side shows the ``skeleton" version of the scene.}
\label{fig:left_lane_change}
\end{figure}

In order to generate training labels for short-term road intentions, we implement a basic path planning system which can perform simple collision avoidance, generate paths for short-term intentions such as lane change, decelerate, accelerate, car follow, and lane keep as shown in Table~\ref{tab:short_term_intents}. 
\begin{table}
\scalebox{1.0}{\parbox{.5\linewidth}{%
\begin{center}
    \begin{tabular}{ | c | l |}
    \hline
    Intention class & Intent name \\
    \hline
    0  &  Lane keep \\
    1  &  Change lane to right \\
    2  &  Change lane to left  \\
    3  &  Decelerate  \\
    4  &  Accelerate  \\
      \hline
    \end{tabular}
\end{center}
}}
\caption{List of the short-term driving intentions in the basic path planning system.}
\label{tab:short_term_intents}
\end{table}
The path for intentions such as lane-keep, car follow, accelerate and decelerate are generated using the car location, longitudinal velocity, accelerations, deceleration, as well as HD map information of the road segment polygons, lane marker polygons, etc., whereas for lane-change intention vehicle's lateral velocity, acceleration, deceleration are also considered. Stopping intention is triggered when either there is obstacle vehicle in the current planned path or traffic light stopping request is presented. Examples of the generated plans are shown in Fig.~\ref{fig:lane_keep}, Fig.~\ref{fig:urban} for intentions such as lane keep, car follow, stop, etc., in Fig.~\ref{fig:left_lane_change} for lane change as the blue dots and arrow.

Once a path is planned for the vehicle, the action controls, maneuvers of the vehicle are implemented using a proportional-integral-differential (PID) controller which locks the car coordinates to the generated paths based upon cross-track-error (CTE), $e(t)$, as shown in Eq.\ref{eq:pid}

\begin{align}
 c(t)&=K_p e(t) + K_i \int_0^t e(t^\prime) dt^\prime + K_d \frac{de(t)}{dt}, \nonumber \\
 \label{eq:pid}
 \end{align}
 where the proportional component compensates any discrepancies between current vehicle coordinates and the way points in planned path, and the differential component assumes a constant error rate at infinitesimal time scale allowing for somewhat predictive control capability. There are four mechanical controls of the vehicles including the throttle (gas pedal), brake, hand brake, steering wheel, as well as two electrical controls of signal lights on the right and left hand side, which are all exposed to client programs as either IPC shared memory or TCP/IP communications. We launch two main computing threads, one for reading the world model, vehicle status from maya, performing the basic path planning as well as vehicle controls \footnote{ Reading data takes less than few milliseconds with a data readout rate of 5 Hz, whereas basic path planning as well as controls takes less than 15 milliseconds.}, and the other thread for recording training feature data as well as ground truth training labels of intentions to the TFRecords. The two computing threads communicate through a first-in-first-out (FIFO) queue data structure.

\section{Our Prediction Model} 
 
\begin{figure}[!ht]
  \centering
    \includegraphics[width=0.5\textwidth]{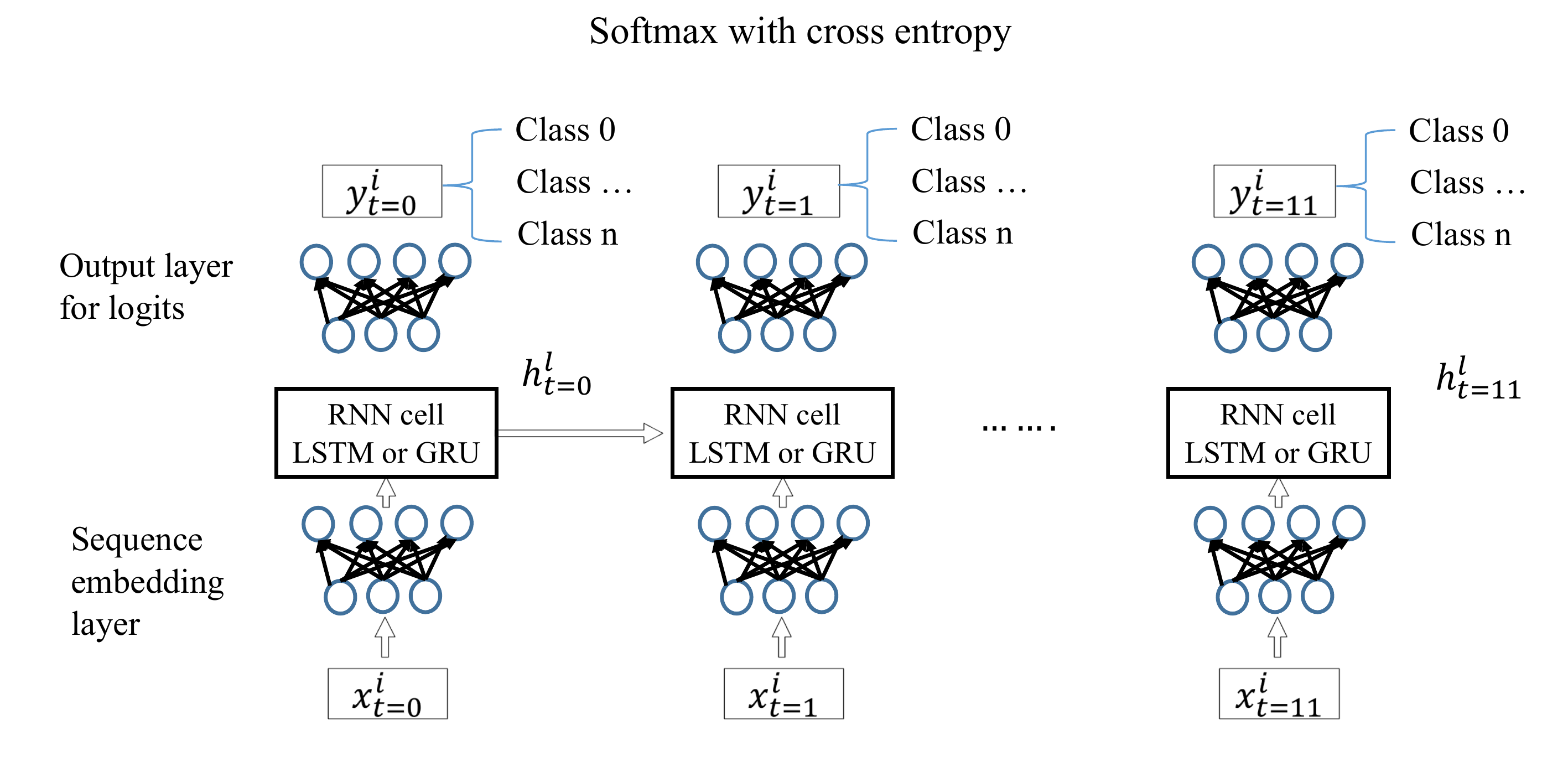}
\caption{The architecture of the seer net where the baseline configuration uses 12 frames to construct a sequence.}
\label{fig:rnn}
\end{figure}

In order to implement a time dependent intention classification model, we reformalize the use of LSTM and GRU from sequence generations to sequence classifications. Subtleties of the goal reside in taking into account the temporal correlations within the sequence, i.e., classifying the intention labels at each time step of a sequence is necessary for a better capture of the intention transitions. As shown in Fig.~\ref{fig:rnn}, we build a standard recurrent neural network, denoted as seer net, that takes measurements of sequence data as input vectors, $x_{t}^i$, at each time step, and outputs a sequence of classes of intentions as predictions. Backward propagation is governed by minimizing the loss function constructed out of predictions and ground truth labels at each time step within the sequecne.
The input vectors are embedded using fully connected layers with a Rectified Linear Unit (ReLu) activation, thereafter fed into the RNN cells of either LSTM or GRU cells. A typical LSTM memory cell has input, output, regularized input and forget gates, as well as the memory cells as shown in Eq.\ref{eq:lstm},

\begin{align}
 i_t&=\sigma\left( W^{(i)}x_t + U^{(i)}h_{t-1} + b^{(i)} \right) \nonumber \\
 o_t&=\sigma\left( W^{(o)}x_t+U^{(o)}h_{t-1}+b^{(o)} \right) \nonumber \\
 f_t&=\sigma\left( W^{(f)}x_t+U^{(f)}h_{t-1}+b^{(f)} \right) \nonumber \\
 g_t&=\tanh\left( W^{(g)}x_t+U^{(g)}h_{t-1}+b^{(g)} \right) \nonumber \\
 c_t &= i_t  \odot g _t + f_t \odot c_{t-1} \nonumber \\
 h_t &= o_t \odot \tanh(c_t),
 \label{eq:lstm}
 \end{align}
where the out put gates of LSTM are used to generate ``logits", $y_{t}^i$, thus class pseudo probabilities using softmax function with cross entropy loss. We have tried various drop out rate for the LSTM cells and in the end we use no drop out as baseline model. The output layers of the cells are wired into another fully connected network to generate logits, and the loss function is then computed using the softmax function with cross entropies on the labels of intentions at each time step. We also experiment with GRU cells,
\begin{align}
 z_t&=\sigma\left( W^{(z)}x_t + U^{(z)}s_{t-1} \right) \nonumber \\
 r_t&=\sigma\left( W^{(r)}x_t+U^{(r)}s_{t-1} \right) \nonumber \\
 h_t&=\tanh\left( W^{(h)}x_t+U^{(h)}(s_{t-1} \odot r_t) \right) \nonumber \\
 s_t &= (1 - z_t)  \odot h _t + z_t \odot s_{t-1} \nonumber \\
 \label{eq:gru}
 \end{align}, where there is less parameters to learn and fewer gates to propagate in recurrence, and the cells are directly exposed as output layer for classifications. The performances of the different cell types are tabulated in Sec.~\ref{sec:quantitative_measurement}. The baseline configuration for the LSTM or GRU cell latent dimensionality is set to 128 units.

\section{Precision and recall measurements} 
\label{sec:quantitative_measurement}

We perform a cross validation measurement to evaluate, in a quantitative manner,  the precision of the intention predictions, per time step, using different recurrent structures of LSTM and GRU cells, where training and testing samples are generated using the same basic path planner and vehicle PID controllers, but with different random seeding. There is some ``contamination" between intention of car following and intentions of for example accelerate, decelerate, as sometimes car follow mode can also accelerate and decelerate as well to maintain a constant distance between ego vehicle and the vehicle in front, thus we drop any sequences that contain a label of intention of car follow as a data cleaning procedure. The precision and recall measurements are tabulated in Table~\ref{tab:precion_recall} where for each of the four datasets we measure how many of the time steps that are predicted to belong to a certain intention actually agree with the ground truth labels, as well as how many of time steps that are labeled with the ground truth intentions get recognized by the predictions. 

\begin{table}
\scalebox{0.90}{\parbox{.5\linewidth}{%
\begin{center}
    \begin{tabular}{ | c | l | l | l | l | }
    \hline
Intent    & \multicolumn{2}{|c|}{Precision [$\%$]} & \multicolumn{2}{|c|}{Recall [$\%$]} \\
index    &		LSTM  	&  GRU  & LSTM  	&  GRU\\
\hline
\multicolumn{5}{|c|}{Dataset A}  \\
\multicolumn{5}{|c|}{only use lateral and longitude coordinates}  \\
\hline
 0 &  $97.11\pm0.03$  &  $96.27 \pm 0.04$  & $98.42\pm0.02$ & $98.06\pm0.03$ \\
 1 &  $89.08\pm0.20$&  $90.69 \pm 0.19$ &   $85.55\pm0.22$& $79.32\pm0.25$\\
 2 &  $93.66\pm0.16$&   $87.30 \pm 0.21$&   $84.53\pm0.22$& $81.62\pm0.24$\\
 \hline
 \multicolumn{5}{|c|}{lateral and longitude coordinates, car heading angle}  \\
\hline
 0 &   $99.36 \pm 0.02$ & $99.33 \pm 0.02$ &  $98.36\pm0.02$ &  $98.92\pm0.02$\\
 1 &   $92.24 \pm 0.16$ & $94.43 \pm 0.14$ & $96.73\pm0.11$  &  $96.63\pm0.11$\\
 2 &   $91.65 \pm 0.16$ & $94.44 \pm 0.14$ &  $96.46\pm0.11$ &  $96.14\pm0.12$\\
\hline

 \multicolumn{5}{|c|}{Dataset B} \\
  \multicolumn{5}{|c|}{only use lateral and longitude coordinates}  \\
    \hline
  0 &  $96.03 \pm 0.04$  &  $95.93\pm0.04$  & $99.08\pm0.02$ & $98.86\pm0.02$ \\
 1 &  $92.35 \pm 0.17$&  $93.32\pm0.17$ &   $79.17\pm0.25$& $78.08\pm0.25$\\
 2 &  $96.41 \pm 0.13$&   $92.64\pm0.18$&   $78.85\pm0.25$& $78.63\pm0.25$\\
   \hline
 \multicolumn{5}{|c|}{lateral and longitude coordinates, car heading angle}  \\
  \hline
 0  &  $99.21 \pm 0.02$ & $99.30 \pm 0.02$ &  $99.16 \pm 0.02$ & $98.65 \pm 0.02$ \\
 1  &  $95.10 \pm 0.13$ & $89.44 \pm 0.18$ &  $96.19 \pm 0.12$ & $96.41 \pm 0.11$ \\
 2  & $95.81 \pm 0.12$ & $96.86 \pm 0.11$  &  $95.17 \pm 0.13$ & $95.72 \pm 0.12$ \\
 \hline
  \multicolumn{5}{|c|}{Dataset C} \\
   \multicolumn{5}{|c|}{lateral and longitude coordinates, car heading angle}  \\
  \hline
 0  &  $99.20\pm0.01$ & $99.01\pm0.01$ &  $98.07\pm0.02$ & $99.10\pm0.01$ \\
 1  &  $84.28\pm0.17$ & $95.37\pm0.10$ &  $94.35\pm0.11$ & $92.33\pm0.13$ \\
 2  & $90.69\pm0.15$ & $90.96\pm0.15$  &  $94.10\pm0.13$ & $93.00\pm0.14$ \\
 \hline

  \multicolumn{5}{|c|}{Dataset D} \\
   \multicolumn{5}{|c|}{lateral and longitude coordinates, car heading angle}  \\
  \hline
 0  &  $98.73 \pm 0.03$ & $98.84\pm0.03$ &  $97.93 \pm 0.03$ & $94.98\pm0.05$ \\
 1  &  $88.08 \pm 0.27$ & $77.04\pm0.33$ &  $88.26 \pm 0.27$ & $89.90\pm0.25$ \\
 2  & $83.32 \pm 0.32$ & $66.15\pm0.36$  &  $93.99 \pm 0.22$ & $94.24\pm0.21$ \\
 \hline

    \end{tabular}
\end{center}
}}
\caption{We perform a cross validation measurement to evaluate the precision of the intent predictions where training and test samples are generated using the same basic path planner and vehicle controllers but with different random seeding. There is some ``contamination" between intention of car follow and intention of accelerate and decelerate as sometimes car follow mode can accelerate and decelerate as well to maintain a constant distance from the vehicle in front. The uncertainties here are only statistical assuming a binomial distribution of precision measurement.}
\label{tab:precion_recall}
\end{table}
Results for different sequence length, different cell types, different input data features, different scenes are also compared in the table. The uncertainties here are only statistical error assuming a binomial distribution of precision or recall measurement. In general, the per-time-step precision of predictions are precise with little misclassifications, and we choose to use an overall prediction result averaged over all time steps as the sequence prediction in order to cancel some of the ``noisy" predictions caused by per-time-step misclassifications.

 \section{Demonstration of the inference performances}
 \label{sec:qualitative_measurement}

\begin{figure}[!ht]
  \centering
    \includegraphics[width=0.5\textwidth]{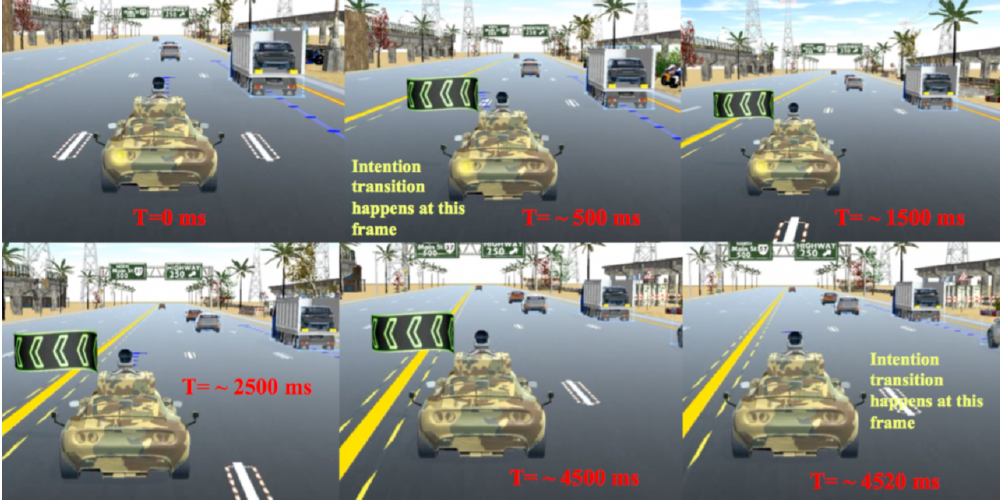}
\caption{Demonstration of the prediction results is shown here, where $T=0 \text{ ms}$ frame (top left sub-figure) is the time when the driver initially signals for changing to the left lane. About 500 ms later, which correspond to about 3 time steps at read out rate of 5 $Hz$, the recurrent neural network built on top of a sequence of 12 time steps predicts an intention transition from lane keep to changing to the left lane. In total, it takes about 4.5 second to fulfill a lane-change intention as shown in the bottom right sub-figure.}
\label{fig:demo_lane_change}
\end{figure}

 An example of prediction of the lane-change intention is analyzed qualitatively in great detail in Fig.~\ref{fig:demo_lane_change}, where $T=0 \text{ ms}$ frame (top left sub-figure) is the time when the driver initially signals for changing to the left lane. About 500 ms later, which correspond to about 3 time steps at read-out rate of 5 Hz, the recurrent neural network built on top of a sequence of 12 time steps predicts intention transition from lane keep to changing to the left lane. Here the predicted intention for the whole sequence is determined by counting the intention predictions at each time step, where the majority intention among the 12 individual predictions is selected to stand for the intention of the whole sequence. We use majority voting approach, instead of the predicted result at every time step, to determine the sequence intention mainly to avoid being influenced by the per frame prediction errors. The green moving arrow sign stands for prediction result of changing to the left lane as compared to the text format of intentions shown earlier in this article, which are not prediction results but the ground truth labels for training purposes. The blue dots and arrow stands for the trajectory or path prediction which is generated based upon the intention prediction as well as vehicle kinetics at current time step.
 
In total, it takes about 4.5 second to fulfill a lane-change intention as shown in the bottom right sub-figure where predicted intention transitions from changing to the left lane back to lane keep. We do not use any information of the signal light in the process of prediction, only the vehicle coordinate, trajectory information is used for prediction.

\begin{figure}[!ht]
  \centering
    \includegraphics[width=0.5\textwidth]{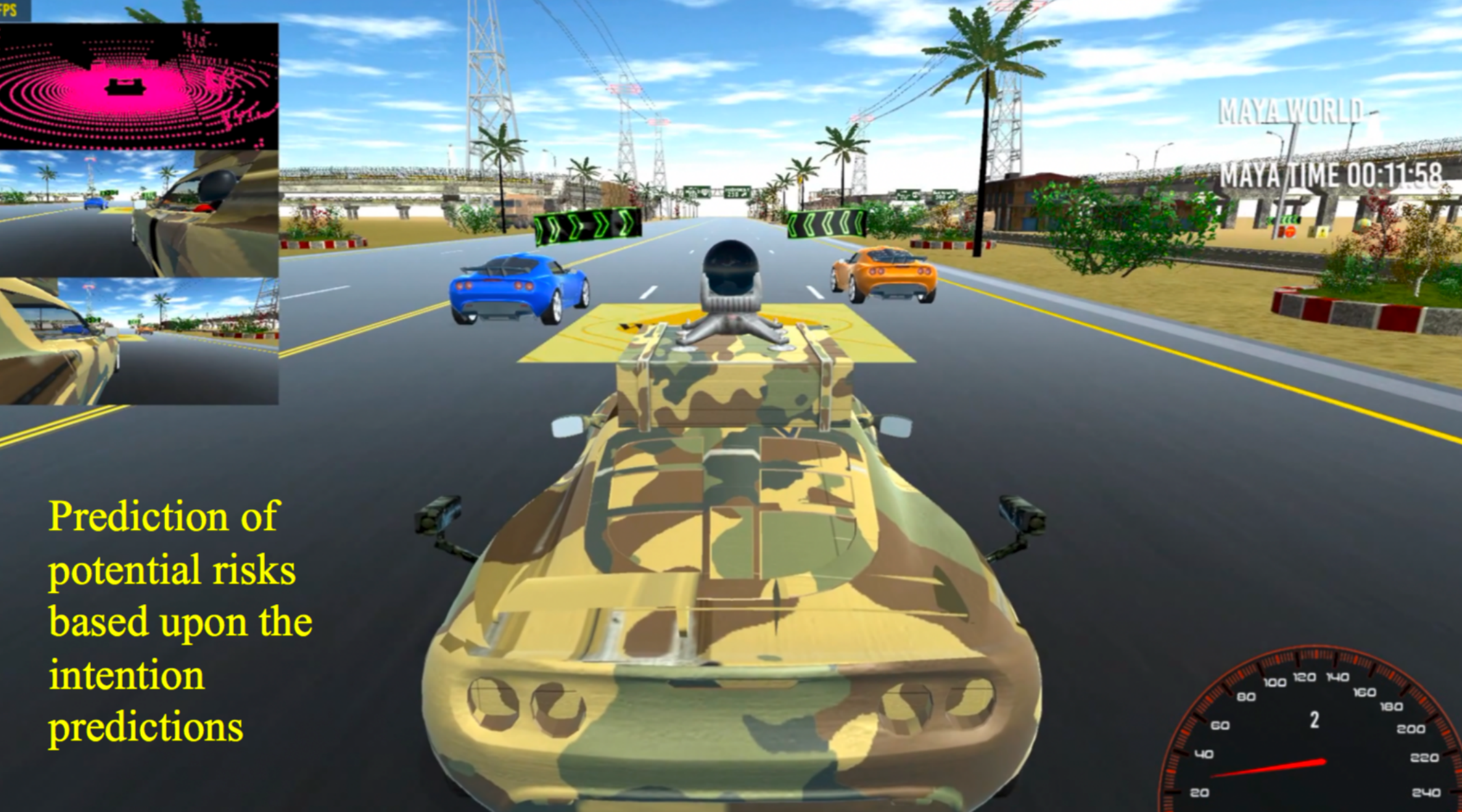}
\caption{Demonstration of the prediction of potential risks is shown where a conflict of interests on a spot between two vehicles are predicted based upon the intention prediction results. The risky zone or spot is marked by the yellow warning box on the road surface.}
\label{fig:demo_risks}
\end{figure}

 Prediction of some potential road risks can be performed based upon intention prediction results, a demonstration of such risk is shown in Fig.~\ref{fig:demo_risks} where two vehicles are predicted to claim the same spot in the road. The risky zone or spot is marked using the yellow warning box on the road. Many other road risks can be identified and derived from fundamental intention predictions.
 
All the inference processes are implemented using Tensorflow C API where a typical inference batch that contains about 10 vehicles takes less than 10 milliseconds using a GeForce GTX 1080 GPU card.

\section{Conclusions}

Driving intention predictions are crucial for autonomous driving vehicles to avoid possible risks, collisions, and to drive in a defensive manner. A predictive driving system is proposed and prototyped where a chain of traffic scene generation, scene rendering in 3D experimental simulation environment, data collection and pipeline, training and cross-validation of a recurrent neural network based predictive model, and real-time inference engine are presented in this article. Precise per-frame short-term intention prediction results are measured using the system, while some of the standard scenes are also open to public for research purposes. A few inference examples of intention predictions, risk predictions are also demonstrated in the paper.  Future work involves predictions of relatively long-term intentions such as goals, behaviors, etc., using much longer sequences of data.

\section{Appendices}

The public version of Maya simulator can be found either in AWS S3 bucket or Docker hub \cite{maya_docker}. Some of the input arguments are listed in Table~\ref{tab:maya_arguments} where the simulator can be run either in graphics mode for visualization, or in batch mode for data collection purposes. We plan to open more traffic scenes, especially scenes dedicated for testing challenging traffic scenarios, collision avoidances, etc.

\begin{table}
\scalebox{0.7}{\parbox{.5\linewidth}{%
\begin{center}
    \begin{tabular}{ | l | l | l |}
    \hline
    Parameterization & Description & Default \\
    \hline
    $--log\_frequency$ & vehicle data log frequency $\times$ 10 ms & 20\\
    $--dump\_png$  & send image packet out & False\\
    $--dump\_lidar$ & send LIDAR point cloud & False\\
    $--dump\_hd\_map$ & send HD map out & False \\
    $--dump\_png\_frequency$ & image output frequency $\times$ 10 ms & 500 \\
    $--dump\_lidar\_frequency$ & LIDAR output frequency $\times$ 10 ms & 500 \\
    $--dump\_hd\_map\_frequency$ & HD map output frequency $\times$ 10 ms & 5000 \\
    $--log\_port$ & ipc or socket for vehicle data & N/A \\
    $--log\_port\_image$ & ipc or socket for image data & N/A\\
    $--log\_port\_lidar$ & ipc or socket for LIDAR data & N/A\\
    $--log\_port\_hd\_map$  & ipc or socket for HD map data & N/A\\
    $--control\_port$ & ipc or socket to send control signals & N/A \\
    $--no\_bbox$ & disable bounding box rendering & False\\
    $--no\_vehicle\_name$ & disable vehicle name & False \\
    $--log\_lidar\_frequency$ & frequency to updates & 100 \\
    $$ & LIDAR measurement $\times$ 10 ms &  \\
    $--skeleton$ & switch to ``skeleton" mode & False \\
    $--traction$ & tire traction, tweak for drifting & 0.5 \\
    $--forward\_slip\_limit$ & forward slip limit & 0.1 \\
    $--sideway\_slip\_limit$ & side way slip limit & 0.1 \\
    $--start\_position$ & agent start position and orientation & N/A\\
    $--camera\_distance$ & following camera distance to ego vehicle &  5 m\\
    $--camera\_height$ & following camera height & 1.5 m\\
    $--camera\_look\_at\_height$ & focus spot height & 0.5 m\\
    $--num\_lanes\_segments$ & number of lane segments of high way & 31 \\
    $--road\_width$ & high way road width & 45 m \\
    $--road\_length$ & high way road length &  1000 m \\
    $--lane\_width$ & high way lane width & 6.5 m \\
    $--num\_lane\_right$ & number of lanes on  & 3\\
     & the right-hand-side of high way &  \\
    $--num\_lane\_left$ & number of lanes on  & 3 \\
    & the left-hand-side of high way & \\
    $-batchmode$ & run in batch mode for data collection & False \\
    $-nographics$ & disable graphics & False \\
      \hline
    \end{tabular}
\end{center}
}}
\caption{Some of the configuration parameters for maya simulation are shown where the simulator can be run either in graphics mode for visualization, or in batch mode for data collection purposes.}
\label{tab:maya_arguments}
\end{table}

%
\bibliographystyle{unsrt}

\bibliography{intent_prediction}  
%
%

\end{document}